\documentclass[10pt, conference, letterpaper]{IEEEtran}  

\usepackage[letterpaper,
  left=0.625in, right=0.625in,    
  top=0.75in,                     
  bottom=1.05in                    
]{geometry}

\IEEEoverridecommandlockouts
\usepackage{cite}
\usepackage{amsmath}
\usepackage{amssymb}
\usepackage{bm}
\usepackage{amsmath,amssymb,amsfonts}
\usepackage{algorithm}
\usepackage{algpseudocode}
\usepackage{graphicx}
\usepackage{subfig}
\usepackage{textcomp}
\usepackage{xcolor}
\usepackage[hidelinks]{hyperref}
\usepackage{balance}
\def\BibTeX{{\rm B\kern-.05em{\sc i\kern-.025em b}\kern-.08em
   T\kern-.1667em\lower.7ex\hbox{E}\kern-.125emX}}
\begin{document}

\title{Energy-Efficient Quantized Federated Learning for Resource-constrained IoT devices
}



\author{
\IEEEauthorblockN{Wilfrid Sougrinoma Compaor\'e\IEEEauthorrefmark{1}, Yaya Etiabi\IEEEauthorrefmark{1}, El Mehdi Amhoud\IEEEauthorrefmark{1}, and Mohamad Assaad\IEEEauthorrefmark{2}}
\IEEEauthorblockA{\IEEEauthorrefmark{1}College of Computing, Mohammed VI Polytechnic University, Benguerir, Morocco\\
Email: \{Compaore.sougrinoma-wilfrid, yaya.etiabi, elmehdi.amhoud\}@um6p.ma}
\IEEEauthorblockA{\IEEEauthorrefmark{2}CentraleSup\'elec, University of Paris-Saclay, France\\
Email: mohamad.assaad@centralesupelec.fr}
}

\maketitle

\begin{abstract}
Federated Learning (FL) has emerged as a promising paradigm for enabling collaborative machine learning while preserving data privacy, making it particularly suitable for Internet of Things (IoT) environments. However, resource-constrained IoT devices face significant challenges due to limited energy, unreliable communication channels, and the impracticality of assuming infinite blocklength transmission. 
This paper proposes a federated learning framework for IoT networks that integrates finite blocklength transmission, model quantization, and an error-aware aggregation mechanism to enhance energy efficiency and communication reliability. The framework also optimizes uplink transmission power to balance energy savings and model performance. 
Simulation results demonstrate that the proposed approach significantly reduces energy consumption by up to 75\% compared to a standard FL model, while maintaining robust model accuracy, making it a viable solution for FL in real-world IoT scenarios with constrained resources. This work paves the way for efficient and reliable FL implementations in practical IoT deployments.
\end{abstract}

\section{Introduction} Federated learning (FL) has emerged as a promising machine learning paradigm for distributed systems where privacy and data locality are paramount. Instead of transferring raw data to a central server, FL enables numerous clients, such as Internet of Things (IoT) devices, to collaboratively train a global model using local data\cite{lim2020federated}.This approach mitigates privacy concerns by keeping data on devices, but introduces significant communication and energy efficiency challenges, especially critical for resource-constrained IoT environments.

IoT devices, including sensors, drones, and low-power computing units, are often limited in both computational and communication resources, restricting their capacity for extensive processing or frequent data transmissions. 
The communication bottleneck is one of the main obstacles in FL, as frequent transmission of large model updates from each device to a central server can overwhelm available bandwidth and lead to excessive energy consumption on constrained devices\cite{imteaj2020federated }.
Many FL frameworks \cite{chen2019joint}, \cite{kim2022tradeoff} assume ideal communication conditions, where updates from clients are transmitted with ample bandwidth and minimal errors, an assumption rarely valid in practical IoT deployments where communication channels are unreliable and latency requirements are strict.

To address this bottleneck, prior studies have proposed compression and quantization techniques to reduce communication load. For instance, gradient sparsification, structured updates, and quantization \cite{wu2016quantized} aim to shrink model updates in order to reduce the communication load. Quantization, in particular, reduces data size by lowering precision, thereby decreasing transmission energy and storage demands \cite{chen2023energyefficient}. Other efforts to improve energy efficiency in FL for IoT include frameworks such as \cite{kim2023green}, which optimize FL for low-power devices by reducing resource demands without significantly compromising accuracy.  However, these methods often assume infinite blocklength transmission, ignoring practical IoT constraints like latency bandwidth constraints, and energy limitations.

In reality, finite blocklength transmission (FBT) is more appropriate for IoT networks, as it reflects the trade-offs between data rate, blocklength, and error probability \cite{polyanskiy2010finite}. Initially studied in channel coding theory, FBT has recently gained traction as a practical model for latency-sensitive IoT applications. Studies such as \cite{she2020radio} and \cite{durisi2016toward} highlight its relevance in IoT networks, where unreliable channels and short packet sizes are common.

However, in this mode, ensuring reliable communication with short data packets introduces a non-negligible probability of transmission errors, which can degrade the model aggregation process if not properly addressed. This is especially crucial in FL, where inaccurate updates from clients due to transmission errors can lead to degraded global model performance \cite{she2020radio}. Existing FL frameworks, however, often overlook these potential transmission errors, leading to inefficiencies in both model accuracy and energy use.

In response to these challenges, this paper proposes a novel approach for enhancing energy efficiency in FL for IoT environments by leveraging finite blocklength transmission while explicitly addressing transmission errors. Our approach integrates model quantization not only during local training but also during uplink transmission, and introduces an error-aware aggregation mechanism at the server to adjust for transmission inaccuracies. By taking transmission errors into account and optimizing uplink transmission power, this approach seeks to strike an optimal balance between energy efficiency, model precision and performances, specifically designed to meet the constraints of IoT devices.
\begin{figure}[h!]
   \centering
     \includegraphics[scale=.35]{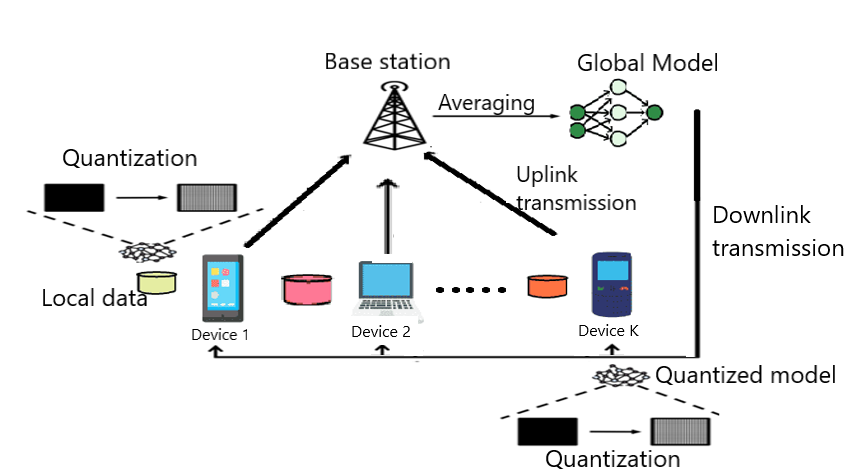} 
     \caption{An illustration of the System Model}
   \label{al}
\end{figure}
To the best of our knowledge, this work is the first to tackle the issue of communication efficiency in FL by incorporating finite blocklength transmission with an error-aware aggregation strategy and uplink transmission power optimization. Through simulations, we demonstrate that our approach achieves substantial energy savings while maintaining acceptable model performance, making it a viable solution for FL in resource-limited IoT devices.

The rest of the paper is organized as follows: Section II presents the system model, including our proposed quantization strategy. Section III describes the proposed approach for energy optimization. Section IV provides simulation results, and Section V concludes the study and set forth some perspectives.
\section{System model}
In this paper, we consider an FL system consisting of \( N \) devices connected to a base station (BS) as shown in Fig \ref{al}. Each device \( k \) has a local dataset \( \mathcal{D}_k \), consisting of labeled samples { \( \{x_{k,l}, y_{k,l}\} \)  \( l = 1, \ldots |\mathcal{D}_k| \)}. In this dataset, \( x_{k,l} \) is an input features vector  and \( y_{k,l} \) is the corresponding output label. 
Here, the main objective is to collaboratively train a global model $\boldsymbol{w} \in \mathbb{R}^d$ across all devices and the base station by minimizing the global loss function defined as:
\begin{equation}
\min_{\boldsymbol{w}} f(\boldsymbol{w}) = \sum_{k=1}^N \frac{|\mathcal{D}_k|}{\mathcal{D}} f_k(\boldsymbol{w}),
\label{equ1}
\end{equation}
where $f_k(\boldsymbol{w})$ is the local empirical loss at client $k$, defined as:
\begin{equation}
f_k(\boldsymbol{w}) = \frac{1}{|\mathcal{D}_k|} \sum_{l=1}^{|\mathcal{D}_k|} \ell(\boldsymbol{w}, x_{k,l}, y_{k,l}),
\end{equation}
and $\mathcal{D} = \sum_{k=1}^N |\mathcal{D}_k|$ denotes the total dataset size across all clients. The function $\ell(\cdot)$ is typically a non-convex loss (e.g., cross-entropy) computed on sample $(x_{k,l}, y_{k,l})$.

Addressing the optimization problem (\ref{equ1}) generally entails a back-and-forth communication process between the BS and the devices. 
In practical applications such as the IoT networks, devices are frequently limited by energy constraints. These limitations make it impractical for them to execute energy-intensive FL processes.
As a result, it is essential to control the precision of FL to minimize energy consumption in computation, memory access, and data transmission. 
Thus, employing a quantized neural network (QNN) with weights and activations in a fixed-point format, as opposed to the traditional 32-bit floating-point format, becomes essential.
\subsection{Quantized neural networks:}
Quantization reduces the bit-width of weights, biases, and activations, allowing computations with integers instead of 32-bit floats. 
This lowers memory and energy costs, making neural networks more efficient for resource-constrained IoT devices.

In our approach, the weights are first clipped to the interval $[-1, 1]$ to ensure compatibility with the fixed-point format, and quantization is applied both during local training and uplink transmission, optimizing data precision while reducing energy consumption. We employ stochastic quantization to represent values in a fixed-point format \([ \sigma.\omega' ]\), where \( n \) bits encode the entire value: \( \sigma \) represents the integer part (1 bit), and \( \omega' \) the fractional part (\( n - 1 \) bits). The use of 1 bit for the integer part allows representing signed values in the range $[-1, 1)$, assuming symmetric quantization around zero. The quantization gain $G$ ensures the dynamic range of weights fits this interval without overflow. This \( n \)-bit quantization provides a trade-off between precision and energy efficiency, crucial for IoT-based federated learning under strict power and bandwidth constraints.
\subsection{Quantization Procedure}
The quantization method adopted in this paper enhances compatibility with low-power IoT devices while preserving model performance. It consists of three steps:

\textbf{1. Scaling up}: Each original weight \( \boldsymbol{w} \) is scaled by a quantization gain \( G = 2^{(n-1)} \), which amplifies the weight to fit within the representable integer range: 
 $\boldsymbol{w}_Q = \boldsymbol{w} \cdot G$ 

\textbf{2. Stochastic rounding}: \( \boldsymbol{w}_Q \) is rounded to an integer 
 (floor or ceil ) following the  probability function based on its fractional part:
 \[
R(\boldsymbol{w}_Q) =\begin{cases} 
        \lfloor \boldsymbol{w}_Q \rfloor,&\text{with probability } 1-(\boldsymbol{w}_Q -\lfloor \boldsymbol{w}_Q\rfloor) \\
        \lfloor \boldsymbol{w}_Q \rfloor + 1,&\text{with probability }\boldsymbol{w}_Q -\lfloor \boldsymbol{w}_Q \rfloor.
    \end{cases}
\] 

\textbf{3. Scaling down}: After transmission, the quantized value \( \boldsymbol{w}_Q \) is scaled back down by dividing by \( G \) to approximate the original weight \( \boldsymbol{w} \): \hspace{4pt}
          $\boldsymbol{w}_r = \boldsymbol{w}_Q/G$
This approach reduces communication overhead in federated learning while maintaining model accuracy.
\subsection{Federated learning updates aggregation model}
The model is trained using the stochastic gradient descent (SGD) algorithm as follows:
\begin{equation}
\boldsymbol{w}^k \gets \boldsymbol{w}^k - \eta \nabla f_k(\boldsymbol{w}^{Q,k}, \xi_k),
\end{equation}
where \( \eta \) is the learning rate, \( \boldsymbol{w}^{Q,k} \) is the quantized value of \(\boldsymbol{w} \) for device \( k \), and \( \xi_k \) is a mini-batch for the current update. We adopt the FedAvg algorithm\cite{mcmahan2017communication}
for the training process. The entire training process is divided into rounds (global iterations), each consisting of \( I \) local updates at each client. At the beginning of the \( t \)-th round, the BS randomly selects a set of devices \( \mathcal{N}_t \) with \( |\mathcal{N}_t| = K \) and transmits the current global model \( w_t \) to these devices. Each selected device \( k \) quantizes and updates its local model by performing \( I \) steps of SGD on its local loss function as follows:
\begin{equation}
\boldsymbol{w}_{t, i}^k = \boldsymbol{w}_{t, i-1}^k - \eta_t \nabla f_k \left( \boldsymbol{w}_{t, i-1}^{Q,k}, \xi_k^i \right),\forall i = 1,\ldots, I
\end{equation}
where \( \eta_t \) is the learning rate at the \( t \)-th round. After completing \( I \) steps of SGD, each selected client computes the local model update \( \Delta \boldsymbol{w}_{t}^{k} = \boldsymbol{w}_{t,I}^{k} - \boldsymbol{w}_{t,0}^{k} \) and then quantizes the update with the same quantization precision applied to the model for training. We denote the quantized value of \( \Delta \boldsymbol{w}_t^{k} \) as \( \Delta \boldsymbol{w}_t^{Q,k} \).The BS averages the received model updates to generate the next global model as follows:

\textbf{1) Incorporating transmission errors into update aggregation:}
    Unlike the traditional aggregation approach \cite{kim2023green},
    \begin{equation}
        \boldsymbol{w}_{t+1} = \boldsymbol{w}_t + \frac{1}{K} \sum_{k \in \mathcal{N}_t} \Delta \boldsymbol{w}_{t}^{Q,k},
    \end{equation}
our model accounts for transmission errors in client-server communication. Let \( q \) denote the probability of transmission error due to finite blocklength communication. When an error occurs, the update \( \Delta \boldsymbol{w}_t^{Q,k} \) is ignored, leading to:
\[
    \widehat{\Delta \boldsymbol{w}}_t^{Q,k} = \Delta \boldsymbol{w}_t^{Q,k} \cdot \lambda_k,
    \]
    where \( \lambda_k \) is the reliability factor:
    \[
    \lambda_k = 
    \begin{cases}
    1, & \text{successful transmission (probability } 1 - q) \\
    0, & \text{failed transmission (probability } q).
    \end{cases}
    \]
    This ensures that only reliable updates contribute to model aggregation.

\textbf{2) Aggregation formula:}
    To integrate transmission errors, the global model update is
    \begin{equation}
        \boldsymbol{w}_{t+1} = \boldsymbol{w}_t + \frac{\sum_{k \in \mathcal{N}_t} \alpha_k \cdot \widehat{\Delta \boldsymbol{w}}_t^{Q,k}}{\sum_{k \in \mathcal{N}_t} \alpha_k},
    \end{equation}
    where the client weight $\alpha_k = |D_k|/D$.
    Thus, updates from reliable transmissions (\(\lambda_k = 1\)) are fully considered, while those from failed transmissions (\(\lambda_k = 0\)) are ignored, while ensuring that clients with larger datasets have proportionally more influence on the global update.    
The FL system repeats this process until the global loss function converges to a target accuracy constraint $\epsilon$.
\begin{algorithm}
\caption{Quantized FL Algorithm}
\begin{algorithmic}[1]
\State \textbf{Initialization:} $K$, $I$, $\boldsymbol{w}_0$, $t = 0$, target accuracy $\epsilon$
\Repeat
    \State The BS randomly selects a subset of devices $\mathcal{N}_t$ and broadcasts $w_t$ to the selected devices;
    \For{each device $k \in \mathcal{N}_t$}
         \State Quantize $\boldsymbol{w}^{k}_t$ to get $\boldsymbol{w}^{Q,k}_t$
        \State Train $\boldsymbol{w}^{Q,k}_t$ by performing $I$ steps of SGD;
        \State Each device $k \in \mathcal{N}_t$ transmits $\Delta \boldsymbol{w}_{t}^{Q,k}$ to the BS;
    \EndFor
    \State The BS generates a new global model
    \[
        \boldsymbol{w}_{t+1} = \boldsymbol{w}_t + \frac{\sum_{k \in \mathcal{N}_t} \alpha_k \cdot \widehat{\Delta \boldsymbol{w}}_t^{Q,k}}{\sum_{k \in \mathcal{N}_t} \alpha_k}
    \]
    \State $t = t + 1$;
\Until{target accuracy $\epsilon$ is reached } ;
\end{algorithmic}
\end{algorithm}
\subsection{Energy model}
The energy consumption model accounts for the power required for both local model training and data transmission by devices. The energy use of the base station (BS) is excluded, as it typically has a steady energy source. According to the model from \cite{chen2019joint}, the energy consumed by a device  k for training and transmission  at each round can be formulated as :

\textbf{1) Local training energy}\\
The energy consumed for local training is : 
\begin{align}
     {{e^{k,l}(n)} = \beta C f^2 d_n I}
\end{align}
where $\beta$ is energy consumption coefficient of the device, $C$ the number of cycles of the central processing unit, $f$ is clock frequency and $I$ the number of local iterations. $d_n$: amount of information processed per iteration with $d_n = d * n$ where $n$ is the number quantization bits and $ d $ is the number of variables of $w$.  

\textbf{2) Transmission energy}\\
We consider a point-to-point transmission with short packets through a quasi-static fading channel, assuming full channel state information (CSI) for rate adaptation. The fading follows a Rayleigh distribution, with the channel gain \( h \) remaining constant over \( M \) symbols, the blocklength. From  \cite{Shehab2019}  the achievable rate depends on the SNR $\rho$, block length \( M \), and error probability \( q \), and is approximated by :
\begin{align}
r \approx \mathcal{C}(\rho |h|^2) - \sqrt{\frac{V(\rho |h|^2)}{M}} Q^{-1}(q)
\end{align}
where:
\( \mathcal{C}(x) = \log_2(1 + x) \) is Shannon’s channel capacity, \hspace{30 pt}
\( V(x) = \left( 1 - (1 + x)^{-2} \right)(\log_2 e)^2 \) denotes the channel dispersion, indicating capacity variability with SNR,
\( Q(x) = \int_x^\infty \frac{1}{\sqrt{2\pi}} e^{-t^2/2} \, dt \) is the Gaussian Q-function, representing the tail probability of the standard normal distribution.

The energy consumed for the uplink transmission is : 
\begin{align}
     {{e^{k,u}(n)} = \text{$\tau$} \times P_{tx}= \frac{d_{n}^u}{B_k.r} \times P_{tx}}
\end{align}
where $\tau$ is the transmission time, $B_k$ is the uplink bandwidth (Hz), $P_{tx}$ is the transmission power, $d_{n}^u = d^u * n$, $n$ is the number of quantization bits, and $ d^u $ is the number of the parameters of the quantized model to transmit.
\section{Proposed approach for energy optimization}
We present an energy minimization problem that also ensures convergence to a specified accuracy level. The total energy consumption in our FL system is given as follows:\\
\begin{equation}
e(n) = \sum_{t=1}^{T} \sum_{k \in \mathcal{N}_t} e^{k,l}(n) + e^{k,u}(n) 
 \end{equation}
where T is the total number of communications rounds.\\
Our objective is to minimize the expected total energy consumption in the FL process. For that, we have to find the optimal values of the number of quantization bits $n^*$, the tolerable transmission error probability $q^*$ and the transmit power $p_{tx}^*$ until convergence under the target accuracy $\epsilon$ and it can be formulated as follows: 
\begin{align}
    e_{n^*,p_{tx}^*,q^*} &= \min_{n,p_{tx},q}\mathbb{E}(e(n)) \\
    &= \min_{n, p_{tx}, q} \mathbb{E} \left[ \sum_{t=1}^{T} \sum_{k \in \mathcal{N}_t} e^{k,l}(n) + e^{k,u}(n) \right] 
    \label{eq:mon-equation}
\end{align}
\[\text{s.t.} \quad n \leq n_{\max}, \quad \mathbb{E}[f(\bm{w}_T)] - f(\bm{w}^*) \leq \epsilon 
\]
\[
\tau_{pr} \leq  \tau_{limit}
\]

where $\mathbb{E}[f(\boldsymbol{w}_T)]$  denotes the expected value of the global loss function after \( T \) global iterations, \( f(\boldsymbol{w}^*) \) is the minimum value of the global loss function \( f \), $\tau_{pr}$ is the expected maximum time per round,  which consists of the local computation and the uplink transmission time for each device and $\tau_{limit}$ is the time constraint per round.

According to \cite{kim2022tradeoff}, given that \( K \) devices are randomly chosen out of N at each global iteration, we can express the expected value of the objective function in $e_{n^*, p^*,q*}$ as follows:
\begin{align}
    f_e(n) &= \mathbb{E} \left[  \sum_{t=1}^{T} \sum_{k \in \mathcal{N}_t} e^{k,l}(n) + e^{k,u}(n)  \right] \\
    &= \frac{K T}{N} \sum_{k=1}^{N} \left( e^{k,l}(n) + e^{k,u}(n) \right).
\end{align}
By finding the minimum value of T denoted as $T^*$ that ensures the convergence. We can also minimize the total time $\tau_{total}$ as follows :
\[
\text{$\tau_{total}$} = T^* \cdot \text{$\tau_{pr}$}
\]  It can be defined as:
\[
\text{$\tau_{pr}$} = \mathbb{E} \left[ \max_{k \in \mathcal{N}_t} \left( \tau^{u}_k + \tau^{comp}_k \right) \right]=\frac{K }{N} \sum_{k=1}^{N} \left( \tau^{u}_k + \tau^{comp}_k\right)
\] 
\[\text{$\tau_{pr}$}  =\frac{K }{N} \sum_{k=1}^{N} \frac{d^u n}{B_k\cdot r_{k} } + \frac{\text{$MacOps/iteration$}}{\text{$C_{comp}$} }  \cdot I
\] where $\tau^{u}_k$ and $ \tau^{comp}_k$ are respectively the uplink transmission and local computation time.The computation capacity \( C_{comp} \) represents the processing power of a device, typically measured in FLOPs(Floating point operations per Second), and determines the speed at which local updates are computed.

To relate \( T \) to \( \epsilon \), we made some assumptions fairly standard and widely used in the convergence analysis of the well established FedAvg algorithm. We assume that the loss function is \( L \)-smooth and \(\mu\)-strongly convex, with the variance and squared norm of the stochastic gradient respectively bounded by \( \sigma_k^2 \) and \( H \) for each device \( k \in \mathcal{N}_t\).\\
We analyze the convergence rate with packet drop model, where \( q \) represents the probability that a packet (i.e., a gradient update) is dropped due to transmission errors.
From the works in \cite{li2020convergence} and \cite{Zheng2020Design}, the convergence bound for \\  \( \Delta_t = \mathbb{E}\left[ \|\bm{w}_t - \bm{w}^*\|^2 \right] \) is given by:
\begin{equation}
\Delta_{t+1} \leq (1 - \eta_t \mu) \Delta_t + \eta_t^2 E,
\end{equation}
where
    \( \eta_t \) is a diminishing step size,
    \( \mu \) is a strong convexity constant,
    \( E \) is a bound on the variance due to gradient noise and is given by:\\
\begin{small}
\begin{equation}
E = \sum_{k=1}^{N} \frac{\sigma_k^2}{N^2} + 6L\Gamma 
+ (8(I - 1)^2 + \frac{4(N - K)I^2}{K(N - 1)} ) H^2 
+ \frac{4d I^2 m^2}{K \left(2^n - 1\right)^2} 
\end{equation}
\end{small}
where  \( \Gamma \) is the degree of non-I.I.d and $m \geq 0$. 

The objective is to prove that \( \Delta_t \leq \frac{v}{t + \gamma} \) where \( v \) and \( \gamma \) depend on the system parameters, including the packet drop rate \( q \).
\subsection*{\textbf{Incorporating packet drop rate \( q \):}}
With packet drops, the probability of successful transmission is \( 1 - q \), meaning only a fraction \( 1 - q \) of the updates are successfully received. One can show easily (derivations are skipped for brevity) that the convergence bound is modified to:
\begin{equation}
\Delta_{t+1} \leq (1 - \eta_t \mu (1 - q)) \Delta_t + \eta_t^2 \frac{E}{1 - q}.
\end{equation}
Here:
\begin{itemize}
    \item \( \eta_t \mu (1 - q) \) represents the effective convergence rate after accounting for packet drops,
    \item \( \frac{E}{1 - q} \) reflects the increased variance due to packet drops.
\end{itemize}

\subsection*{\textbf{Choice of diminishing step size \( \eta_t \):}}

To achieve \( \Delta_t \leq \frac{v}{t + \gamma} \), we choose a diminishing step size: 
    $\eta_t = \frac{\beta}{t + \gamma}$ ,
where \( \beta \) and \( \gamma \) are parameters to be determined.

\subsection*{\textbf{Modified convergence bound:}}

Substituting \( \eta_t = \frac{\beta}{t + \gamma} \) into the convergence bound:
\begin{align}
\Delta_{t+1} &\leq \left(1 - \frac{\beta \mu (1 - q)}{t + \gamma}\right) \Delta_t + \frac{\beta^2 E}{(t + \gamma)^2 (1 - q)}.
\end{align}
In fact, using induction, we assume that \( \Delta_t \leq \frac{v}{t + \gamma} \) and show that it holds for \( \Delta_{t+1} \). We choose \( \beta = \frac{2}{\mu} \), then we select:
\[
v = \max\left(\frac{4E}{(1 - q) \mu^2}, (\gamma + 1) \Delta_1\right),
\]
\[
\gamma = \max\left(I, \frac{8L}{(1 - q) \mu}\right) - 1.
\]
With these values, we ensure that \( \Delta_t \leq \frac{v}{t + \gamma} \) holds.
Then by the strong convexity of \( f(\cdot) \) and from \ref{eq:mon-equation}, we have:
\begin{equation}
\quad \mathbb{E}[f(\bm{w}_T)] - f(\bm{w}^*)  \leq \frac{L}{2} \Delta_t \leq \frac{L}{2} \frac{v}{\gamma + t} \leq \epsilon.
\end{equation}
Therefore, the required number of iterations \( T \) to achieve convergence is: \hspace{10pt}      $T ={Lv}/{2\epsilon} - \gamma.$ 

Thus, the total time and energy to reach convergence are multiplied by \( T \).
Consequently, our optimization problem becomes:
\begin{align}
    \min_{n, p_{tx}, q} \left[\frac{K}{N} \left(\frac{Lv}{2\epsilon} - \gamma\right) \sum_{k=1}^{N} \left( e^{k,l}(n) + e^{k,u}(n) \right) \right]
\end{align}
subject to the constraint on time per round:
\[
\frac{K }{N}  \left(\sum_{k=1}^{N} \frac{d^u n}{B_k\cdot r_{k} } + \frac{\text{$MacOps_{/iteration}$}}{\text{$C_{comp}$}}  \cdot I  \right) \leq \tau_{limit}
\]
To solve this problem, we adopt the covariance matrix adaptation evolution strategy (CMA-ES) which is a derivative-free optimization algorithm designed for non-convex, high-dimensional problems. It adapts a covariance matrix to guide the search efficiently. In our approach, CMA-ES optimizes \( P_{\text{tx}} \) and \( q \), ensuring energy-efficient learning while maintaining communication reliability. The algorithm iteratively samples solutions, evaluates them, and updates its distribution to improve convergence.

\section{Simulation results}
For our simulations, we consider a FL setup with a total of \( N = 100 \) devices, where \( K = 10 \) devices randomly selected each round. The MNIST dataset \cite{lecun1998gradient} is used for training. Although MNIST is a simple dataset, it is widely used in FL and quantization literature to benchmark new algorithms due to its light computational demands and availability of baselines. Future work will include experiments on more complex datasets such as CIFAR-10 or FEMNIST to demonstrate scalability. Unless stated otherwise, the key system parameters are set as follows: the bandwidth for each device \( B_k = 10 \) MHz, power spectral density of white noise \( N_0 = -100 \)dBm, number of iterations \( I =3\), target accuracy \( \epsilon = 0.1 \), \( \beta = 10^{-27}\) J/cycle,  CPU frequency \( f = 1 \) GHz, \( C = 40\) cycles, computation capacity \( 3.7\times 10^{12} \)FLOPs, and \( L = 0.097 \), \( \mu = 1\), \( m = 0.01\), \( H = 0.25\), \( \sigma_k^2 = 0.001\), \( \Gamma = 0.6\), \( \Delta_1 = 0.01\), and \( M = 1000\) symbols. The learning rate of SGD is set to 0.001.  The time constraint  $\tau_{limit}$ is 1 second maximum per round.

We implement a QNN composed of two quantized convolutional layers: the first with 32 kernels of size \( 3 \times 3 \) and the second with 64 kernels of the same configuration, both using a padding of 1 and a stride of 1. Each convolutional layer is followed by a ReLU activation and a \( 2 \times 2 \) max pooling. The model also includes two quantized fully connected layers, where the first layer has 128 units. In this setup, the model requires 4,241,152 MAC operations and has a total of 421,642 weights. The MAC count was computed using standard formulas for convolutional and fully connected layers, based on input/output dimensions and kernel sizes.  

\begin{figure}[!t]
    \centering
    \subfloat[Transmitted power evolution]{
        \includegraphics[width=0.47\linewidth]{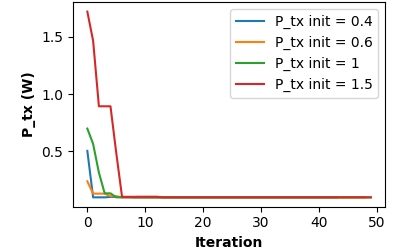}
        \label{subfig:power}
    }
    \subfloat[Transmission error evolution]{
        \includegraphics[width=0.47\linewidth]{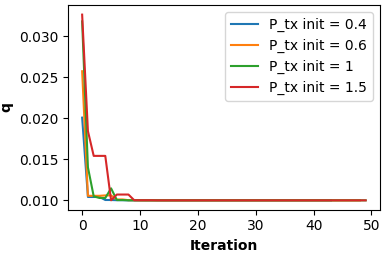}
        \label{subfig:error}
    } 
    
    \subfloat[Objective function value]{
        \includegraphics[width=0.47\linewidth]{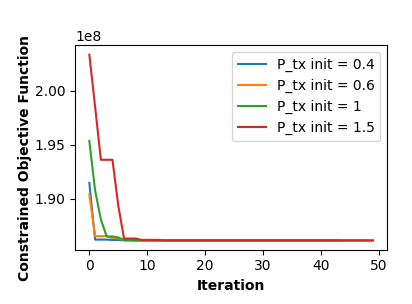}
        \label{subfig:objective}
    }
    \subfloat[Constraint satisfaction]{
        \includegraphics[width=0.47\linewidth]{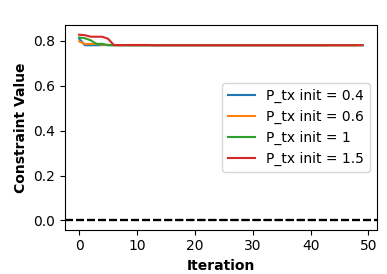}
        \label{subfig:constraint}
    }
    \caption{Convergence of CMA-ES for different initial \( P_{\text{tx}} \) values, showing rapid optimization of \( P_{\text{tx}} \), \( q \), and the objective function while maintaining constraint satisfaction. Each subfigure represents a key metric in the optimization process.}
    \label{fig:Optimization}
\end{figure}

\begin{figure}[h!]
    \centering
    \subfloat[Training accuracy]{
        \includegraphics[width=0.45\linewidth]{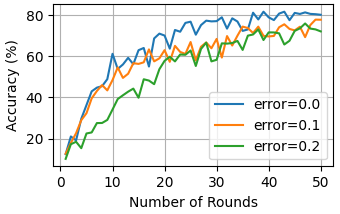}
        \label{subfig:train_acc}
    }
    \subfloat[Validation accuracy]{
        \includegraphics[width=0.45\linewidth]{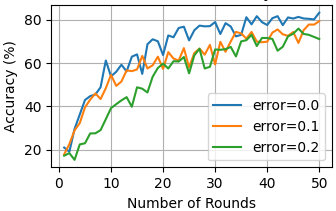}
        \label{subfig:val_acc}
    }
    
    \subfloat[Training loss]{
        \includegraphics[width=0.45\linewidth]{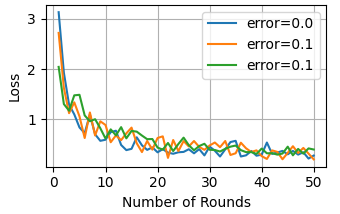}
        \label{subfig:train_loss}
    }
    \subfloat[Validation loss]{
        \includegraphics[width=0.45\linewidth]{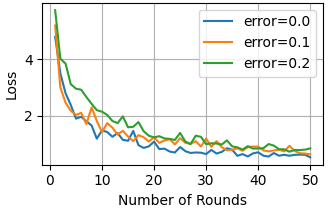}
        \label{subfig:val_loss}
    }
    \caption{Impact of transmission errors on federated learning performance.}
    \label{fig:error_analysis}
\end{figure}

To solve our optimization problem, we first optimize \( P_{\text{tx}} \) and \( q \) using CMA-ES, conducted within \( P_{\text{tx}} \in [0.1, 2] \) and \( q \in [0.01, 0.99] \). The results in Fig. \ref{fig:Optimization} show a clear convergence towards the optimal values \( P_{\text{tx}} \approx 0.1 \) and \( q \approx 0.01 \). Fig. \ref{subfig:power} illustrates the evolution of \( P_{\text{tx}} \) over iterations, where all initial values rapidly converge to 0.1, demonstrating the stability of the optimization. Similarly, Fig. \ref{subfig:error} shows that \( q \) quickly reduces towards 0.01, indicating a preference for lower transmission errors. The constrained objective function, shown in Fig. \ref{subfig:objective}, decreases significantly in the first few rounds before stabilizing, confirming effective energy minimization while maintaining time constraint satisfaction. Finally, Fig. \ref{subfig:constraint} presents the evolution of the constraint value, which remains satisfied throughout the optimization process.This confirms the validity and effectiveness of the proposed optimization approach. Using these optimal values of \( P_{\text{tx}}\) and \( q \), we will therefore determine the optimal quantization level within the standard floating point (FP) formats.
In Fig. \ref{fig:error_analysis}, we analyze the impact of transmission errors on federated learning performance. Fig. \ref{subfig:train_acc} shows the evolution of training accuracy over rounds, where higher error rates (\( q = 0.1, 0.2 \)) lead to slower convergence and lower final accuracy compared to the error-free case (\( q = 0.0 \)). Similarly, Fig. \ref{subfig:val_acc} illustrates validation accuracy, where a degradation in performance is observed as \( q \) increases.

Figures \ref{subfig:train_loss} and \ref{subfig:val_loss} display training and validation loss, respectively. Higher error probabilities result in delayed loss stabilization, indicating that transmission errors disrupt the learning process and slow down convergence. These results highlight the importance of mitigating transmission errors to maintain model performance in federated learning. Note that quantization was not applied in this experiment.

\begin{figure}[!t]
   \centering
     \includegraphics[scale=.6]{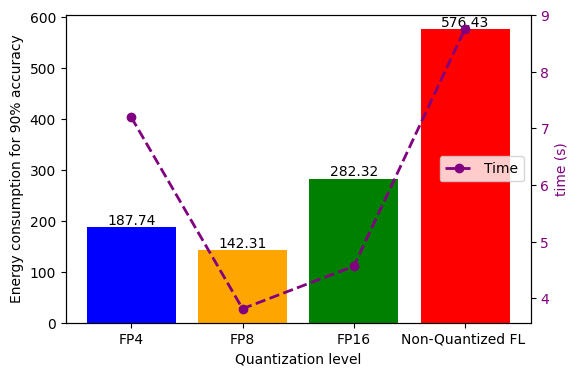}
     \caption{Energy consumption and computation time across quantization levels, showing lower energy use for FP4/FP8 and higher costs for FP16 and Non-quantized FL.}
   \label{tradeOff}
\end{figure}

Subsequently, we evaluated the proposed quantization scheme with error-awareness, that is, we consider an error (\( q \approx 0.01 \)) with different quantization level. In Fig. \ref{tradeOff}, we present energy consumption (bars) and total  time (dashed line) for achieving 90\% accuracy across different quantization levels. Lower-bit quantization (FP4, FP8) significantly reduces energy consumption compared to higher-bit (FP16) and non-quantized FL, but impacts computation time. Notably, FP8 achieves the lowest energy consumption, 75.31\% lower than non-quantized FL, while maintaining a reasonable time overhead, making it the most efficient choice. This demonstrates that our proposed strategy which  jointly optimizes transmission parameters like transmission power (\( P_{\text{tx}} \approx 0.1 \)) and transmission errors (\( q \approx 0.01 \)) along with quantization level, ensures an optimal balance between energy efficiency, computation time, and model performance, demonstrating its effectiveness in constrained FL settings.

\section{Conclusion}
In this paper, we proposed an energy-efficient and communication-aware FL framework for IoT. Using CMA-ES, we optimized transmission power and error probability, achieving rapid convergence while ensuring time constraint satisfaction. Simulations showed that quantization significantly reduces energy consumption, with FP8 achieving 75.31\% lower energy use than standard FL while maintaining efficiency. These findings demonstrate the effectiveness of our approach and provide a foundation for optimizing FL in resource-constrained IoT deployments. These findings also revealed that transmission errors degrade FL performance, highlighting the need for further mitigation strategies.
In future work, an extensive comparison with recent energy-aware FL methods will be conducted to further validate the competitiveness of our approach.
\balance
\bibliographystyle{ieeetr}
\bibliography{references}
\end{document}